**Karolina Rudnicka[1]**

**University of Gdańsk**

**In order that – a data driven study of symptoms and causes of obsolescence**

**Abstract**

The paper is an empirical case study of grammatical obsolescence in progress. The main studied variable is the purpose subordinator *in order that*, which is shown to be steadily decreasing in the frequency of use starting from the beginning of the twentieth century. This work applies a data-driven approach for the investigation and description of obsolescence, recently developed by the Rudnicka (2019). The methodology combines philological analysis with statistical methods used on data acquired from mega-corpora. Moving from the description of possible symptoms of obsolescence to different causes for it, the paper aims at presenting a comprehensive account of the studied phenomenon.

Interestingly, a very significant role in the decline of *in order that* can be ascribed to the so-called higher-order processes, understood as processes influencing the constructional level from above. Two kinds of higher-order processes are shown to play an important role, namely i) an externally-motivated higher-order process exemplified by the drastic socio-cultural changes of the 19th and 20th centuries; ii) an internally-motivated higher-order processes instantiated by the rise of the *to*-infinitive (rise of infinite clauses).

**Keywords:** *in order that*, obsolescence, decline, language change, higher-order processes

[1] https://orcid.org/0000-0001-8097-6086
University of Gdańsk, Faculty of Languages
karolina.rudnicka@ug.edu.pl

## 1. Introduction

This is an empirical paper presenting i) the practical application of a data-driven approach to the investigation of obsolescence recently developed by Rudnicka (2019)[2]; ii) a discussion of possible causes for obsolescence of *in order that* based on the results of an in-depth quantitative and qualitative corpus analysis and insights from the literature. This article sees obsolescence as a process in which a given construction shows a (rapid or gradual, but always relatively consistent) decrease in frequency of use which continues until it disappears, or there are just residual forms left (for a detailed theoretical discussion see e.g. Rudnicka forthcoming).

There are at least three empirically verifiable symptoms of obsolescence: i) negative correlation between time and the frequency of use (e.g. Rudnicka 2019); ii) distributional fragmentation – an increasing restriction of a construction to a certain genre or register (e.g. Leech et al., 2009); iii) paradigmatic atrophy – restriction to only some of the originally available morphological forms or syntactic environments (e.g. Hundt 2014). As for the causes of obsolescence, the literature mentions two – competition and higher-order processes. Subscribing to Rudnicka's (2019) approach to the investigation of obsolescence, the paper argues that competition on the constructional level is never separable from higher-order processes.

The present work is structured as follows: Section 2 presents the purpose subordinator *in order that*; Section 3 deals with symptoms of obsolescence; Section 4 focuses on the causes for the observed obsolescence in progress by looking at developments which might instantiate competition and by relating the frequency changes visible on the constructional level to higher-order changes[3] operating in the language. Section 5 contains a summary of conclusions.

This article takes a construction grammatical perspective, and the phrases investigated are referred to as constructions[4].

---

[2] In (Rudnicka 2019), the focus is on a development of a general framework describing grammatical obsolescence. The network of purpose subordinators is used to demonstrate the framework in action. Here, the *in order that* construction and its obsolescence take the central place.

[3] Higher-order changes are understood as larger trends and processes which concern a "higher level of grammatical organization than the construction" (Hilpert 2013: 14) and which influence the constructional layer of the language "from above".

[4] According to Fried (2015: 974) constructions can be defined as "conventionalized clusters of features (syntactic, prosodic, pragmatic, semantic, textual, etc.) that recur as further indivisible associations between form and meaning."

## 2. A few words on *in order that*

*In order that*, see (1), entered the English language between the seventeenth and eighteenth century[5]. Together with purposive *so* (*that*) and *lest* (an avertive construction), it belongs to purpose subordinators introducing finite clauses.

(1) Playgroups offer the possibility of bringing together mothers and children, <u>in order that</u> they shall both have support and stimulation. (COHA: 1975; NF: The Playground Movement)

*In order that* is used to refer to different-subject situations – the subject of the matrix clause differs from the subject of the purpose clause. Other subordinators in this functional niche include purposive *so* (*that*), see (2), *in order for * to*, (3). In contrast, *in order to*, *so as to* and the purposive *to*-infinitive are used for same-subject situations, see (4).

(2) I did that to let you go <u>so that</u> you would go to him. (COHA:  1990; FIC: Shadow of a Man)

(3) I just want to know how many voices have to speak up <u>in order for</u> someone <u>to</u> get well in this hospital. (COHA: 2003; FIC: If My Mother's Soul Can Hear)

(4) He spent large sums buying fashionable clothing <u>in order to</u> show off. (COHA: 1971; FIC: Passions of the Mind)

Fig. 1 presents the decrease in the frequency of use displayed by *in order that* in the course of the nineteenth and twentieth century. The figure shows that starting from approximately 1910, the frequency of use of *in order that* has begun to decrease in a very steep way. The data have been extracted from the Corpus of Historical American English[6]  (COHA).

---

[5] The construction became a part of the language "in the 18th century or later" (Kortmann 1997: 300) or more precisely "in the first half of the 18th century" (Łęcki & Nykiel 2017: 249). According to OED Online its first attestation took place already in the seventeenth century (s.v. *in order that*, retrieved on September 17, 2020 from http://www.oed.com).

[6] The Corpus of Historical American English (COHA): 400 million words, 1810-2009. Available online at https://corpus.byu.edu/coha/.

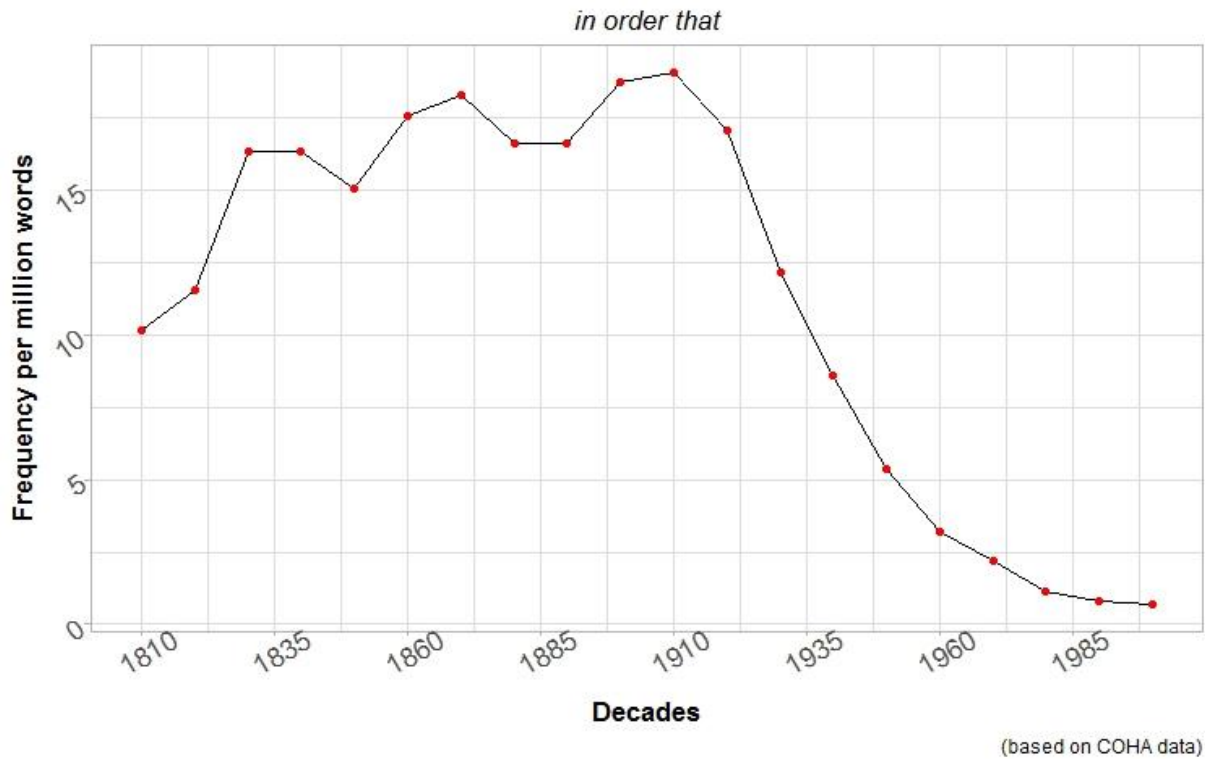

Fig.1: *In order that* – frequency of use in the nineteenth and twentieth century.

### 3. Symptoms of obsolescence

This section investigates *in order that* in the context of symptoms, the presence of which might mean that a given construction is obsolescent. Each of the three bullet points is devoted to one of the symptoms already listed in Section 1 of the present work. For a detailed theoretical discussion of the symptoms, see Rudnicka (forthcoming).

● *Negative correlation between time and the frequency of use*

According to Rudnicka (2019: 29) the first symptom and the necessary condition for obsolescence is the presence of negative correlation between the time and frequency of use. The frequency profile shown in plot in Fig. 1 already suggests that this condition might be fulfilled in the case of *in order that*.

The analysis of data is conducted in R with the use of cor.test function. The method of choice is Kendall's tau test. It takes into account COHA frequency data from the time period of 1900 – 2000. The result we get is tau = -0.9636364 (p-value = 5.511e-07), which means that we detect the presence of very high statistically significant negative correlation between time and the frequency of use. The result corresponds to the plot in Fig. 1. The presence of the negative correlation is to be understood as "the later the decade, the less instances of a given construction are there".

● *Distributional fragmentation*

The increasing concentration of the (remaining) instances of a construction in certain (often formal) genres and registers is frequently mentioned in works dealing with the "negative end" of change (e.g. Hopper and Traugott 2003: 172; Leech et al., 2009; Kempf forthcoming). Nykiel (2014: 9) observes that *in order that* "only ekes out an existence in formal registers after 1900", a claim, which Rudnicka (2019: 131) verified. According to the results obtained from the Corpus of Contemporary American English[7], which are visualized in Fig. 2, even though the most formal genre of the corpus – *academic texts* dominates the picture, instances of *in order that* can be found in every genre represented. 11.6 % of the instances of *in order that* can be found in the section representing *spoken* language, much of which is, in general, considered as less formal than written language (e.g. Reilly and Wren 2003 :97; Jaffe 2012: 202), see (5).

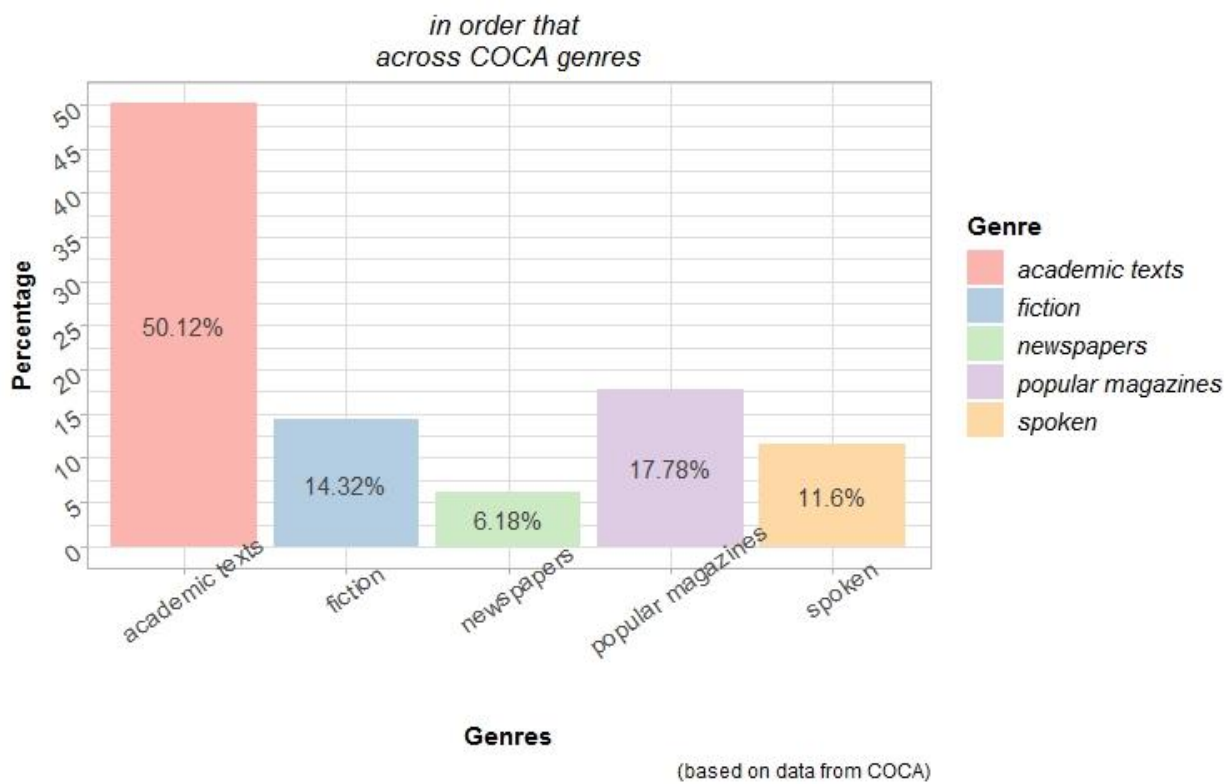

Fig. 2: Distribution of cases of *in order that* across COCA genres (based on Rudnicka 2019: 131).

---

[7] The Corpus of Contemporary American English contains 560 million words of text representing the years 1990 through 2017. It is available online at https://corpus.byu.edu/coca/.

(5) And I can't imagine Mara saying that, well, it's a zig <u>in order that</u> he can zag into a free trader. (COCA: 2009; SPOK: Fox Baier)

To summarize, COCA data shows that more than a half of all the instances of *in order that* can be found in the academic genre. This, however, does not address the question of whether we see an "increasing restriction" – which might be a symptom of obsolescence. One way to deal with this problem is to investigate diachronic frequency data from a multi-genre corpus. COHA is the corpus which could serve this purpose, especially given its size and the fact that obsolescent constructions tend to get rarer across time. Still, the most intuitive and simple comparison of the frequency per million words-curves for each genre will not be conducted, as i) the partition of COHA into genres changes in time, and, ii) *in order that* is, in the second half of the studied period, already very rare, which can bias the results. To illustrate the former point raised, in the decades of 1890 (1990): 55% (47%) of all content is represented by the genre *fiction*; 7% (14%) by *newspaper*; 23% (26%) by *magazine*; 15% (11%) by *non-fiction*. Moreover, differences in the genres' sizes within a given decade introduce an additional bias promoting *fiction* as being the main source of data.

These difficulties can be handled at once by extrapolating the frequencies "as if each of the four genres (in each decade) actually made up 25% of the corpus" (Rudnicka 2019: 136). To this end, we multiply every frequency by 25% and divide it by the corpus's share, the relevant genre had in a considered decade (e.g. 55% for *fiction* in the decade 1890). Fig. 3 presents the extrapolated results.

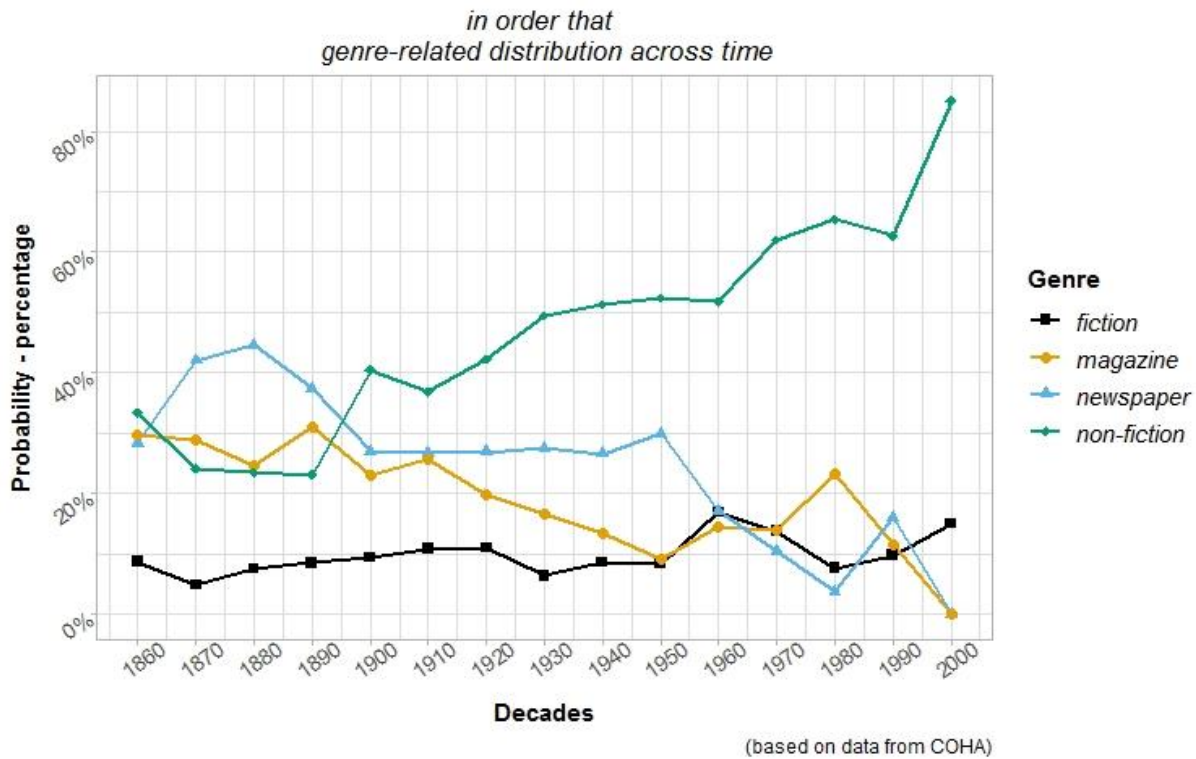

Fig. 3: *In order that* – probability of occurrence in different COHA genres (based on Rudnicka 2019: 141).

As can be seen, there is a rapid increase in the probability of coming across *in order that* in the genre *non-fiction*, at the same time there is a decrease for *magazine* and *newspaper*. For *fiction* the situation seems stable. This picture suggests that we indeed observe an increased concentration of *in order that* in the *non-fiction* genre (also proving that the bias towards *fiction* is mitigated by the extrapolation), a genre which could be seen as the most formal[8] and least colloquial out of the four.

● *Paradigmatic atrophy*

Two examples of paradigmatic atrophy given in the literature are i) an observation pertaining to the modal *shall* – "now almost completely restricted to first person subjects" (Traugott & Trousdale 2013: 67; Leech et al. 2009); ii) "the increasing rarity of the negative contraction in –*n't* with some (though not all) modals" (Leech et al. 2009: 81). Among the properties, which can be tested for to determine whether there are signs of paradigmatic attrition, Rudnicka (2019:

---

[8] *Non-fiction books* is treated as the most formal genre of COHA, as it contains texts which instantiate formal language (according to Leech & Svartnik 2013: 30) such as academic writing (e.g. articles from different peer-reviewed journals), informative books, and official reports.

155) lists 1) the functional alternations depending on the position (initial vs. final)[9] of subordinators in a sentence, see (6) and (7); and 2) the frequency trend concerning their negative forms (such as e.g. *in order that … not*), see (8). However, according to the examination of data extracted from COHA, no signs of paradigmatic atrophy with regard to *in order that* have been detected (Rudnicka 2019: 174).

(6) Soon after he was old enough to walk he was compelled to find work <u>in order that</u> he might eat. (COHA: 1920; FIC: Poor White)

(7) <u>In order that</u> this might happen quickly, they must govern themselves and not make it necessary for Lanny to summon American soldiers to keep order. (COHA: 1948; FIC: Town with the Funny Name)

(8) At night she was to have the company of a woman friend <u>in order that</u> she might <u>not</u> feel lonely, and the following evening I was to be at home again. (COHA: 1910; FIC: The Trail of a Sourdough Life in Alaska)

## 3.1 Symptoms of obsolescence: conclusions

Two out of three possible symptoms are detected for *in order that*, including the necessary condition for obsolescence – the presence of negative correlation. This suggests the construction might indeed be obsolescent.

## 4. Competition and higher-order processes

According to Lass (1997), the equivalence of function is one of the prerequisites for the existence of competition. Section 2 already mentions two potential competitors of *in order that*, namely *in order for * to* and *so that*, which do exist in the same functional niche understood after Lindsay and Aronoff (2013: 135) as "a clearly defined subdomain within its potential domain — a subsystem that is therefore distinct and predictable to a speaker".

Another condition is, according to Hundt & Leech (2012: 176), the presence of a direct reflection in the frequencies of use of the two competing constructions. To check if this is the case, the frequency gains and losses across time can be compared.

---

[9] An adverbial clause of purpose in an *initial position* (a preposed clasue) typically introduces a new, global topic (Thompson 1985), whereas in a *final position* (a postponed clause) it is more likely to introduce a local topic (Ford 1993: 14). Schmidtke-Bode (2009: 126) ascribes "a thematic role in the ongoing discourse" to the preposed clauses and "focal, rhematic function of motivating the event in the matrix clause" to the postponed ones.

**4.1 Methodology:** *in order that* vs. *in order for * to* and *so that*

For *in order that* and *in order for * to*, the frequency of use per million words is simply extracted from the COHA corpus. For *so that*, this will not work, as the construction can be used in both clauses of purpose and result. Still, as Schmidtke-Bode (2009: 30) claims, a purpose clause introduced by *so that* will usually contain a modal auxiliary, which is not the case for clauses of result, compare (9) and (10).

(9)     The doctors gave us this room <u>so that</u> we could have privacy. (COHA: 2009; FIC: Lady Jasmine)

(10)    Immediately the material crumbled in his hands, as though rotting, <u>so that</u> it left only the small, red pebble intact. (COHA: 1951; FIC: The Blind Spot)

To get a high precision and recall, three searches are conducted, namely "so that * _vm*", "so that * * _vm*" and "so that * * * _vm*" . The "*" symbolize the number of words between *so that* and the modal verb ("_vm"). In the next step, frequency gains or losses on a decade-to-decade basis are calculated. Manual spot-checks were conducted to test whether the sentences were indeed purposive uses of *so that* and the precision was assessed as high.

**4.2 Results**

Table 1 contains a sum-up of results for the three constructions – *in order that*, *in order for * to* and purposive *so that* – the sum for the three variants (with one, two and three words between *so that* and the modal verb) which were searched for – in the time period of 1900 – 2000, in which the frequency of use of *in order that* decreases dramatically (see Fig. 1). The purposive *so that* shows a total frequency loss of 24 instances per million words, however, the whole development, compared to the clearly negative trend exemplified by *in order that*, might look like frequency fluctuations. Still, high statistically significant negative correlation between time and the frequency of use (tau = -0.6; p-value = 0.009) confirms the decline. Interestingly, out of the three constructions, only *in order for * to* displays a slight increase in the frequency of use. This increase, however, is definitely too small to account for the decrease shown by *in order that*, so the condition proposed by Hundt & Leech (2012: 176) is not fulfilled for any of the constructions. This result could either mean that the function fulfilled by *in order that* is no longer in the system, or, that the "most intuitive" competitors are also being replaced by other constructions.

| Decade | Frequency gains and losses; from decade to decade; per million words | | |
|---|---|---|---|
| | *in order that* | *in order for * to* | *so that (purposive context)* |
| **1900-1910** | +0.33 | -0.02 | 9.62 |
| **1910-1920** | -2 | +0.43 | -6.8 |
| **1920-1930** | -4.92 | -0.6 | 6.59 |
| **1930-1940** | -3.53 | +0.25 | 2.74 |
| **1940-1950** | -3.28 | +0.6 | -10.09 |
| **1950-1960** | -2.13 | +0.45 | -3.57 |
| **1960-1970** | -1.03 | -0.61 | 2.05 |
| **1970-1980** | -1.03 | +0.2 | -10.55 |
| **1980-1990** | -0.36 | +0.84 | -5.32 |
| **1990-2000** | -0.11 | -0.53 | -8.69 |
| **In total** | **-18.06** | **+1.01** | **-24.02** |

Table 1: Frequency gains and losses for in order that and its natural competitors – *for * to*-infinitive and purposive *so that*.

**4.3 Alternative competitors**: *for * to*-infinitives and *so * modal verb

One characteristic shared by all the three variants in Table 1 is their elaborateness – they might be perceived as lengthy and possibly more formal than their shorter variants, exemplified by infinitival clauses with notional subject introduced by *for* – *for * to*-infinitives and *so * modal verb (without *that*). These constructions can definitely be used with a purposive meaning, however, the fact that this is not always the case, makes data extraction a bit more complicated, compare (11) with (12) for *for * to*-infinitives; and (13) with (14) for *so*.

> (11)     I laid them on the counter <u>for him to see</u>. (COHA: 2001; FIC: My My Repeater)
>
> (12)     She went on without waiting <u>for me to reply</u> which was fortunate. (COHA: 1970; FIC: Secret Woman)
>
> (13)     I'm ruining my hands <u>so you can</u> sit here and watch TV and smoke cigarettes. (COHA: 2000; FIC: Fishing on Sunday)
>
> (14)     And I am tired, too, <u>so I 'll</u> say good-night and go to bed. (COHA: 1970; FIC:

Bay of Noon)

According to De Smet (2013: 79), the very first function of *for * to*-infinitive was that of a purpose adjunct. De Smet further argues that starting from there, this pattern began to spread across different syntactic functions such as the relative clause, noun complement, verb complement and adjunct of comparison. Corpus data show that indeed the frequency of use of *for * to*-infinitive is on the rise, see Fig. 4. What we see confirms the increase in frequency and popularity of *for * to*-infinitives observed by Mair & Leech (2006: 336) and described in detail by De Smet (2013: 79).

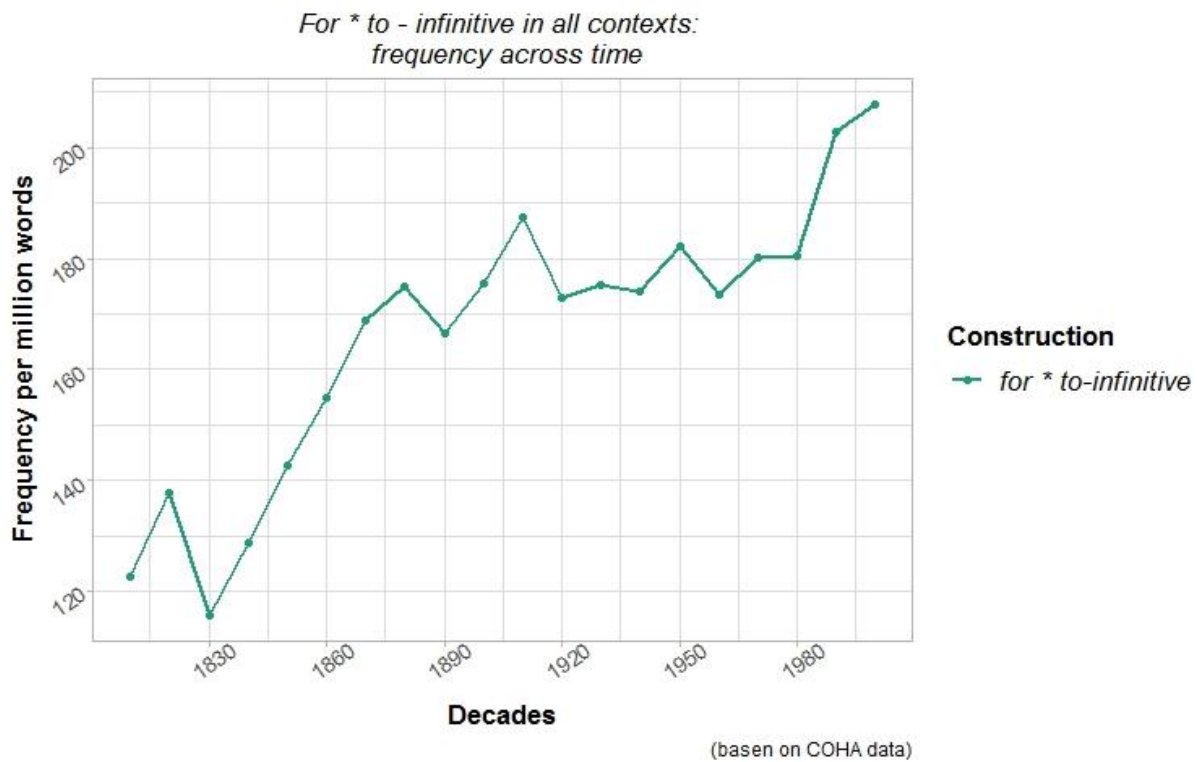

Fig. 4: *For * to*-infinitive in all contexts: frequency across time.

The plot encompasses all the contexts in which the *for * to*-infinitive can be used, still, if we compare the frequencies shown (>200 instances per million words) with the frequency of *in order that*, which was less than 20 instances per million words around 1910 (see Fig. 1), it seems very plausible that at *for * to*-infinitive can at least to some extent account for loss in the popularity of *in order that*.

When it comes to the other shorter variant – *so* in the purposive meaning – the differentiation between purposive, e.g. (13), and non-purposive, see (14), meanings, is not as easy as in the case of the full variants *so that*. Rudnicka (2019: 109) looked at this problem and, in order to determine any frequency trend concerning the usage of purposive *so*, conducted a sampling of 100 random instances of *so* * modal verb ("so * _vm*") from COHA at three points in time – 1900, 1950 and 2000. The sampled instances were manually assessed and classified as purposive or non-purposive. The results in Table 2 show a tendency towards an increase in the number of purposive usages across time. The numbers are large enough to assume that at least some of the frequency decrease of *in order that* and purposive *so that*, might have been caused by the growing popularity of the shorter variant. Figure 5 shows a frequency plot of *in order that*, *in order for* * to*, purposive *so that* and estimated frequency of purposive *so*.

| Decade | *So* * + finite clause Frequency per million words | |
|---|---|---|
| | Purposive semantics | Non-purposive semantics |
| **1900** | 13.87 | 46.45 |
| **1950** | 35.99 | 56.29 |
| **2000** | 73.88 | 80.04 |

Table: 2: Normalised frequency of purposive and non-purposive *so* * + finite clause in 1900, 1950 and 2000 in a sample of 100 per decade (from Rudnicka 2019: 210).

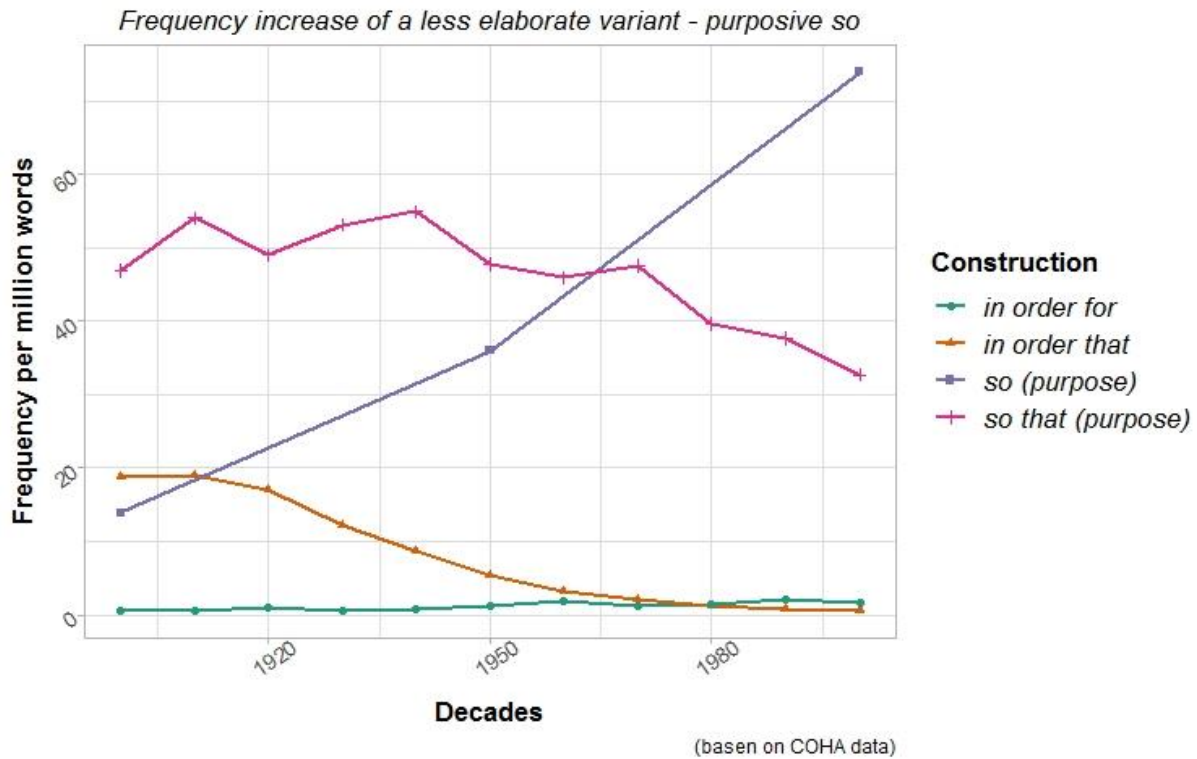

Fig. 5: Frequency across time: *in order that*, *in order for * to*, purposive *so that* and estimated frequency of purposive *so*.

## 4.4 Competition – summary of conclusions

Both prerequisites for competition listed at the beginning of this section, namely the equivalence of function (Lass 1997) and the presence of a direct reflection in the frequencies of use of the two competing constructions (Hundt & Leech 2012: 176), are fulfilled for two constructions:

● purposive *so* (see Fig. 5 for the estimated frequency of the purposive use)

● purposive *for * to*-infinitive (see Fig. 4 for the frequency of use for all contexts)

For the third one *in order for * to*-infinitive, the slight increase visible in Fig. 4 has started later than the decrease of *in order that* and the reflection of the potential competition is barely visible. Purposive *so that* displays a decrease in the frequency of use itself, so the second prerequisite is not fulfilled.

## 4.5 Competition as sign of higher-order processes

Crucially, the variants increasing in the frequency of use share a few characteristics – they are shorter, less elaborate and less formal than the variants displaying a decrease in the frequency of use. This common denominator might suggest that the trend in question is part of a larger change happening in the language. One of such trends might be colloquialization (Mair 1997) – a trend in which the gap between the written and spoken norm narrows (tendency towards more speech-like writing). Other trends include democratization (e.g. recently described by Hiltunen and Loureiro-Porto 2020) often understood as "the removal of inequalities and asymmetries in the discursive and linguistic rights, obligations and prestige of groups of people" (Fairclough 1992: 201) and tendency towards informality (Leech et al. 2009; Farrelly and Seoane 2012: 393). Biber and Finegan (1989) observe that "the development of popular literacy fostered a shift towards more oral styles". Rudnicka (2019: 190) agrees with this and suggests that looking "above" the processes such as colloquialization, we see socio-cultural changes such as the achievement of mass literacy by the American society, the invention of new printing technologies (Hames and Rae 1996: 227) or the introduction of cheap postage in Britain, the so-called Penny Post, in 1840 (Smitterberg 2012). Rudnicka (2019: 182) suggests the name *externally*-motivated higher-order processes for drastic social changes of this kind. The label stems from the terminology of Hickey (2012: 42), according to whom, changes which are triggered and guided by social factors can be seen as *externally*-motivated changes. Table 3 presents a hierarchical account of externally-motivated changes which lead to the situation in which *in order that* becomes obsolescent. In the view of this hierarchy, the constructional competition is only to be seen as a manifestation of what is happening on all the higher levels. The drastic socio-cultural changes "delegate work"[10] to discourse-pragmatic language-change processes influencing speakers' choices and the way constructional level looks like.

| Highest level: *Externally*-motivated higher-order processes | Drastic socio-cultural changes of the nineteenth and twentieth century ● advent of mass-circulation newspapers; ● achievement of mass-literacy by the American society; ● invention and development of new printing technologies etc. |
| --- | --- |
| Higher-level: Discourse-pragmatic language-change processes | Colloquialization; informalization of the language of the media; democratization |

[10] Metaphorically, the socio-cultural changes give the job of exerting influence on the language users to the discourse-pragmatic language-change processes; *delegate* understood after Cambridge Online Dictionary (s.v. *delegate*, retrieved on December 4, 2020).

| **Supra-constructional level** | ● diffusional spreading of shorter and less formal *for * to*-infinitives across different syntactic functions (De Smet 2013); ● general decrease of sentence length observed and tendency towards shorter sentences (Rudnicka 2018) |
|---|---|
| **Constructional level** | ● obsolescence of *in order that*; ● decrease in the frequency of purposive *so that* |

Table 3: The influence of an externally-motivated higher-order process on all the levels down to the constructional level.

Kempf (forthcoming), who investigates loss of the relative particle *so* in early stages of New High German, also observes a hierarchical structure of different causes for the investigated loss and writes "there is not only a multitude of factors but, crucially, a hierarchy of influential factors". The socio-linguistic developments she lists include democratization, the Enlightenment, and literalization. Interestingly, mass readership seems to have played a different role in German than it did in English.

Still, obsolescence of *in order that* could also be associated with changes independent of sociolinguistic factors. Hickey (2012: 42) refers to them as *internally*-motivated changes. By analogy and with regard to higher-order processes Rudnicka (2019: 182), uses the label *internally*-motivated higher-order processes. An example of such higher-order might be the rise of the *to*-infinitve, which was described in detail by Los (2005). The growing popularity of the *for * to*-infinitive very likely results from the still-ongoing process in which *that* clauses are being replaced by infinitival clauses. This process might also explain the emergence of *in order for * to* as an alternative for *in order that* (and possibly as a more elaborate version of the purposive *for * to*-infinitive). The latter construction has still been fully functional when the former came into being. The fact that it introduces an infinitival clause is probably not a coincidence. Table 4 contains a hierarchical account an internally-motivated higher-order process which might, to some extent, account for the obsolescence of *in order that*.

| **Highest level: *Internally*-motivated higher-order processes** | **The rise of the *to*-infinitive** |
|---|---|

| Higher-level: | ● increase in the frequency of non-finite clauses ● decrease in the frequency of subjunctive *that*-clauses (Los 2005: 28; Mair & Leech 2006) |
|---|---|
| Supra-constructional level | ● increase in the frequency of use of the *for * to*-infinitives – the so-called diffusional-spreading (De Smet 2013) |
| Constructional level | ● obsolescence of *in order that*; ● decrease in the frequency of use of *so that*; ● the emergence of *in order for * to* as an alternative for the fully functional *in order that* |

Table 4: The influence of an internally-motivated higher-order process on all the levels down to the constructional level.

### 5. Obsolescence of *in order that*

The paper seeks to join the visible and statistically provable frequency changes on the constructional level (exemplified by *in order that* or *in order for * to*) with larger (but still statistically traceable) trends such as a) decrease of word length of an English sentence; b) growing popularity of *for * to*-infinitives; or c) increase in frequency of "less formal" means of communication. These supra-constructional trends are described as being manifestations of e.g. discourse-pragmatic language change processes such as colloquialization or informalization; or general trends such as decrease in the frequency of subjunctive *that*-clauses. Such higher-level processes are, in turn, hypothesized to take their origin from either 1) drastic socio-cultural changes, referred to as externally-motivated higher-order processes, as exemplified in Table 3; or from 2) major language-internal processes, referred to as internally-motivated higher-order processes, as presented in Table 4. The frequency changes visible on the constructional level can therefore be seen as a resultant of the particular trends and processes working on different levels of language change. The findings of this work suggest that higher-order processes operating at a given time in language, might play an important role in "programming" the future of a construction.

The history of *in order that* also seems to fit the description of marginalization (Hansen 2017: 264) – a process which contrary to grammaticalization "does not lead to the rise of an unmarked, highly frequent grammatical operator, but to elements which occupy a peripheral position in the language system". If, as was the case for *in order that*, the construction turns out to be against higher-order processes operating at a given time, chances are it will become obsolete or it will become a case of marginalization.

**Language corpora & Dictionaries**

**Software**